\documentclass[nonacm]{acmart}
\AtBeginDocument{%
  \providecommand\BibTeX{{%
    \normalfont B\kern-0.5em{\scshape i\kern-0.25em b}\kern-0.8em\TeX}}}
\usepackage{graphicx,tabularx,subcaption}
\usepackage{booktabs}
\usepackage{multirow}
\usepackage[capitalise,nameinlink,compress]{cleveref}
\usepackage{xcolor}
\usepackage{comment}
\usepackage[english]{babel}
\usepackage{fontawesome5}
\usepackage{arydshln}
\graphicspath{{images/}}

\copyrightyear{2024} 

\acmConference[Technical Report]{Technical Report}{Oct}{2024}
\begin{document}


\title[Lessons Learned from EXMOS User Studies]{Lessons Learned from EXMOS User Studies: A Technical Report Summarizing Key Takeaways from User Studies Conducted to Evaluate The EXMOS Platform}

\author{Aditya Bhattacharya}
\affiliation{%
  \institution{KU Leuven}
  \city{Leuven}
  \country{Belgium}
}

\author{Simone Stumpf}
\affiliation{%
  \institution{University of Glasgow}
  \city{Glasgow}
  \country{Scotland, UK}
}

\author{Lucija Gosak}
\affiliation{%
  \institution{University of Maribor}
  \city{Maribor}
  \country{Slovenia}
}

\author{Gregor Stiglic}
\affiliation{%
  \institution{University of Maribor}
  \city{Maribor}
  \country{Slovenia}
}

\author{Katrien Verbert}
\affiliation{%
  \institution{KU Leuven}
  \city{Leuven}
  \country{Belgium}
}

\renewcommand{\shortauthors}{Bhattacharya, et al.}

\begin{abstract}
In the realm of interactive machine-learning systems, the provision of explanations serves as a vital aid in the processes of debugging and enhancing prediction models. However, the extent to which various global model-centric and data-centric explanations can effectively assist domain experts in detecting and resolving potential data-related issues for the purpose of model improvement has remained largely unexplored. This research delves into a comprehensive examination of the impact of global explanations rooted in both data-centric and model-centric perspectives within systems designed to support healthcare experts in optimising machine learning models through both automated and manual data configurations. To empirically investigate these dynamics, we conducted two user studies, comprising quantitative analysis involving a sample size of 70 healthcare experts and qualitative assessments involving 30 healthcare experts. These studies were aimed at illuminating the influence of different explanation types on three key dimensions: trust, understandability, and model improvement. The findings of our investigation underscore a noteworthy revelation: global model-centric explanations alone are insufficient for effectively guiding users during the intricate process of data configuration. In contrast, data-centric explanations exhibited their potential by enhancing the understanding of system changes that occur post-configuration. However, our research shows that the most promising results emerge from a hybrid fusion of both explanation types. This hybrid approach demonstrated the highest level of effectiveness in terms of fostering trust, improving understandability, and facilitating model enhancement among healthcare experts. In light of our study's compelling results, we also present essential design implications for developing interactive machine-learning systems driven by explanations. These insights can guide the creation of more effective systems that empower domain experts to harness the full potential of machine learning in healthcare and other domains. In this technical report, we summarise the key findings of our two user studies.
\end{abstract}




\begin{teaserfigure}
  \centering
  \includegraphics[width=0.82\linewidth]{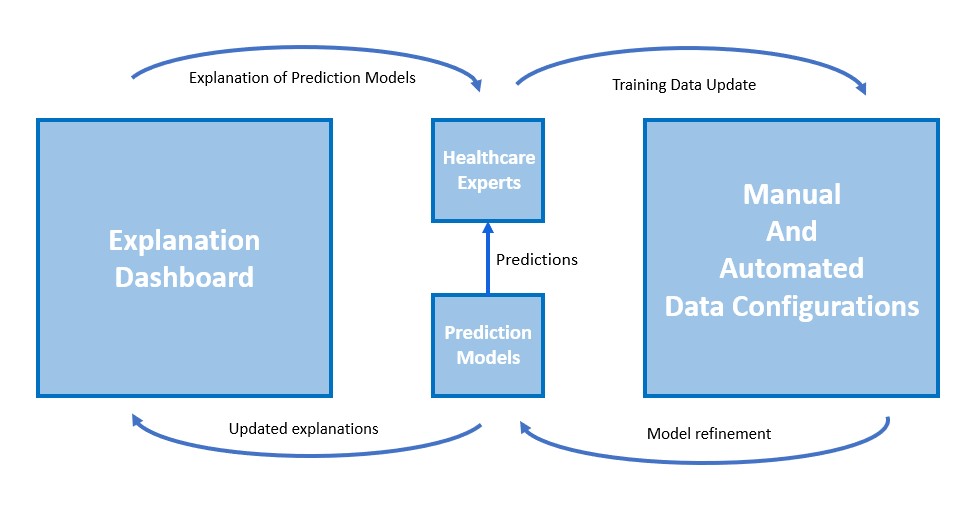}
  \caption{Our EXMOS platform empowers healthcare professionals to optimise predictive models by integrating data-centric and model-centric explanations within Interactive Machine Learning systems through manual and automated data configurations.}
  \Description[Explanatory Model Steering for healthcare]{Our EXMOS platform empowers healthcare professionals to optimise predictive models by integrating data-centric and model-centric explanations within Interactive Machine Learning systems through manual and automated data configurations}
  \label{fig:xil_systems}
\end{teaserfigure}

\maketitle

\section{Introduction}
In recent years, the adoption of artificial intelligence (AI) and machine learning (ML) systems has gained significant traction, particularly in critical domains such as healthcare~\cite{pawar2020incorporating, Caruana2015, Esteva2017-pg}. Central to the effectiveness of these systems is the provision of explanations, a key focus within the field of Explainable AI (XAI). Explanations serve as a means to help end-users develop a clear mental model of these systems, ultimately fostering trust in their operations~\cite{Guidotti2018, adadi2018peeking}.

Diverse types of explanations shed light on various facets of AI/ML systems~\cite{adadi2018peeking, BhattacharyaXAI2022}. There can be global explanations, local explanations, model-centric explanations and data-centric explanations~\cite{Bhattacharya2023, BhattacharyaXAI2022, adadi2018peeking, anik_data-centric_2021}. Moreover, explanations are extremely important for enhancing the understandability of prediction models in interactive machine learning systems (IML)~\cite{fails2003, Amershi_Cakmak_Knox_Kulesza_2014, teso2019, Wang_KDD_2022, teso_leveraging_2022, kulesza_principles_2015, muralidhar2018incorporating, spinner2019explainer, stumpf2009interacting, Guo2022BuildingTI, honeycutt2020soliciting}. 

The concept of Explanatory Interactive Learning (XIL) has emerged when the benefits of XAI and IML are combined by leveraging user feedback via explanations as a human-centric solution for gathering rich end-user feedback to improve AI/ML systems~\cite{teso2019, teso_leveraging_2022, Schramowski2020, bertrand_chi_2023}. Later, researchers urged the necessity of involving domain experts in explanatory interactive systems~\cite {Schramowski2020, lakkaraju2022rethinking}. In domains like healthcare, domain experts with specialised knowledge can leverage their insights to understand complex dynamics within medical data and contribute to model debugging and improvement. For instance, healthcare professionals can interpret the significance of specific medical measurements and their implications for patient outcomes, eventually improving prediction models by modifying the data based on their prior expertise.

Our research \cite{bhattacharya2024exmos, BhattacharyaCHIDC} involved understanding the impact of different types of global explanations in Explanatory Interactive Learning (XIL) systems, particularly within a healthcare context. Our work investigated the effectiveness of data-centric and model-centric global explanations in motivating domain experts to enhance prediction models through two distinct approaches: manual and automated data configuration. Manual configuration empowered users to make informed decisions regarding predictor variables, while automated configuration addressed potential data issues automatically. To explore the influence of explanations on users' choice of data configuration approach, a prototype XIL system was developed, offering three explanation dashboard versions: Data-Centric, Model-Centric, and Hybrid version which combined all explanations from the other two versions. The prototype utilised a Random Forest algorithm on a diabetes prediction dataset and aimed to support healthcare experts in refining models.

Quantitative and qualitative studies involving 70 and 30 healthcare experts were conducted to evaluate the impact of different explanation dashboards on trust, understanding, and model improvement. Through our experiments, we found that participants using the Hybrid explanation dashboard demonstrated significantly better performance in data configuration for model improvement, even though they perceived a higher task load. Notably, the elevated task load did not negatively affect their understanding or trust in the system. Additionally, the study reveals the limitations of global model-centric explanations in guiding users during data configuration compared to data-centric explanations, which were found to be more helpful in comprehending post-configuration system changes due to their holistic nature.

\section{Result Hightlights}
The following are the key takeaways from the two user studies conducted for evaluating our EXMOS platform:
\begin{itemize}
    \item Data-centric explanations were more effective in improving prediction accuracy compared to model-centric explanations, with the hybrid (HYB) approach being the most effective.
    \item HYB users performed better manual data configurations despite a higher perceived task load and longer average hover-time, as more time spent exploring HYB explanations facilitated faster and more effective data configurations.
    \item Model-centric explanation users (MCE) struggled to improve prediction model accuracy in manual configuration, despite spending more time on average.
    \item Lack of data quality information in MCE resulted in less time spent by users to understand automated data corrections.
    \item Findings from our first study showed HYB participants significantly improved prediction model performance compared to Data-Centric Explanation (DCE) participants, with DCE participants slightly outperforming MCE participants.
    \item Combining data-centric and model-centric explanations in healthcare XIL systems was recommended, but initial higher task load was noted. An abstract summary of training data and predictions in the initial view can mitigate this.
    \item Qualitative study participants highlighted the value of data-centric explanations in understanding system changes and suggested including data collection process information for transparency and trust.
    \item Training data visualization on the configuration page improved system comprehension, and local explanations aided in identifying abnormal data records.
\end{itemize}

\section{Tailoring XIL Systems in Healthcare}

Based on our user studies, we propose the following concise recommendations for tailoring XIL systems in healthcare:

\begin{enumerate}
    \item \textbf{Include both Global Data-Centric and Model-Centric Explanations:} Merge both explanation types to empower healthcare experts for effective model steering.
    
    \item \textbf{Include Local Explanations:} Enhance the usefulness, understandability, and actionability of global explanations by incorporating various types of local explanations, including counterfactual and what-if explanations.
    
    \item \textbf{Prioritize Abstraction in Visual Explanations:} Display high-level summary information initially, reserving detailed global and local explanations for subsequent views. Avoid overwhelming users with excessive visualizations.
    
    \item \textbf{Implement Data Filters:} Facilitate user-friendly patient group selection by including data filters in the explanation dashboard and configuration page.
    
    \item \textbf{Present Comprehensive Data Quality Information:} Offer comprehensive data quality details in secondary or tertiary drill-down views, as relevant, particularly for decision-makers and researchers.
    
    \item \textbf{Disclose Data Collection Process:} Provide information on the data collection process in the initial view to clarify the presence of abnormal data values and noisy feature variables.
    
    \item \textbf{Offer Both Manual and Automated Configurations:} Recognize diverse user preferences for control levels by providing options for both manual and automated data configurations, allowing users to override automated settings.
    
    \item \textbf{Importance of Group Feedback and Peer Consensus:} Introduce a peer review and approval system, allowing users to propose changes while ensuring group consensus among healthcare stakeholders.
    
    \item \textbf{Maintain Configuration History to Revert Back to Previous Settings:} Keep a history of data configurations and enable easy revert to previous settings for improved system adoption.
    
\end{enumerate}

\section{Conclusion}
In conclusion, the user studies demonstrated the importance of combining data-centric and model-centric explanations in healthcare XIL systems to enhance model steering and improve prediction model accuracy. This approach provides domain experts with valuable insights and supports more effective data configurations. The design guidelines can act as a blue-print for other researchers and developers for designing and implementing interactive explainability systems.


\bibliographystyle{ACM-Reference-Format}
\bibliography{references}


\end{document}